%% file: main.tex
\documentclass[electronic]{vgtc}             

\graphicspath{{figures/}{pictures/}{images/}{./}} 

\usepackage{times}                     

\usepackage{tabu}                      
\usepackage{booktabs}                  
\usepackage{lipsum}                    
\usepackage{mwe}                       

\usepackage{mathptmx}                  

\onlineid{1275}

\vgtccategory{Research}

\usepackage{comment}
\usepackage{multirow}
\usepackage{amsmath} 
\usepackage{xcolor}

\definecolor{steelblue}{RGB}{70,130,180}
\definecolor{lightblue}{RGB}{173,216,230}

\newcommand{\update}[1]{\textcolor{black}{#1}}

\title{Error Aware Distribution Prediction for Lightweight Implicit Neural Representations}

\author{%
  Zhimin Li\thanks{e-mail: zhimin.li@vanderbilt.edu}\\
  \scriptsize{Vanderbilt University}
  \and Jake D. Balla\thanks{e-mail: jakeballa@arizona.edu}\\
  \scriptsize{University of Arizona}
  \and Joshua A. Levine\thanks{e-mail: josh@cs.arizona.edu}\\
  \scriptsize{University of Arizona}
}

\input{0-abstract}


\teaser{
  \centering
  
\includegraphics[width=\linewidth]{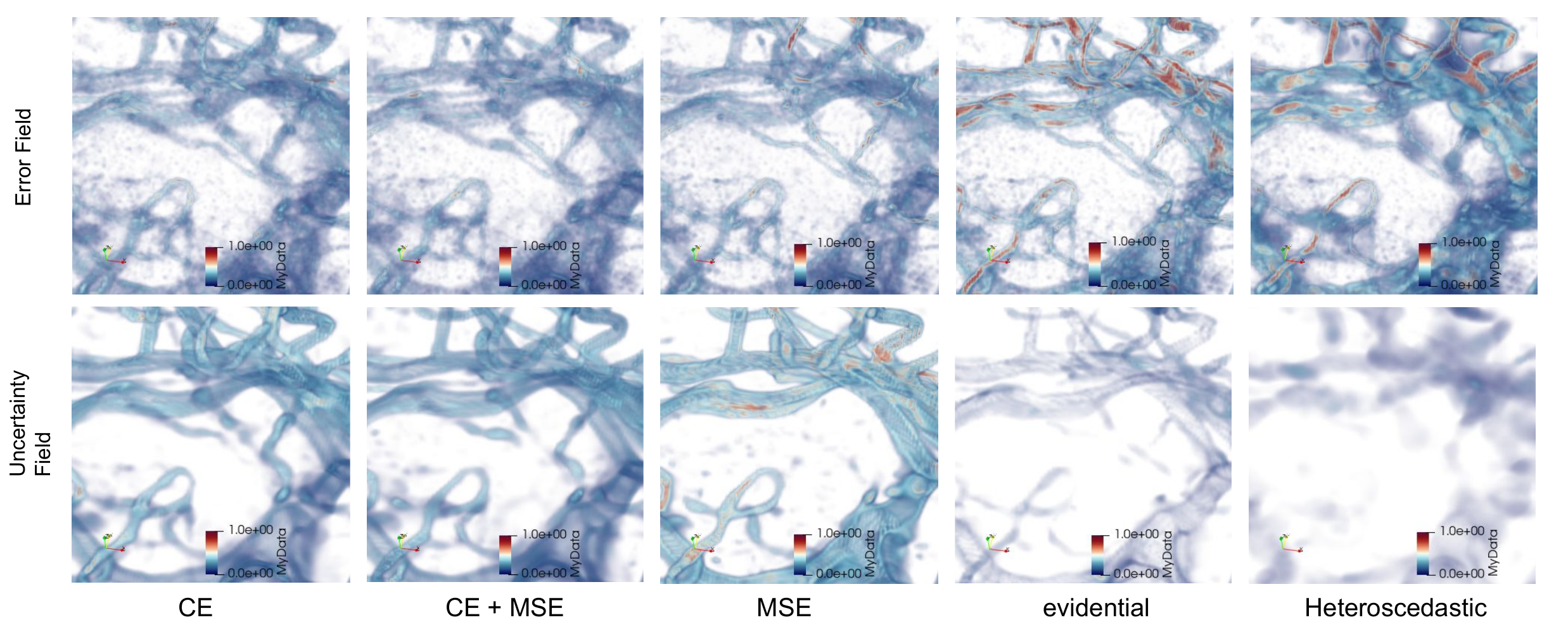}
\caption{Error and uncertainty fields for classification- and regression-based methods on the aneurism dataset. From left to right:cross-entropy(CE), cross-entropy+ mean-squared-error (CE+MSE), mean-squared-error (MSE) in the classification setting, heterocadastic regression, and evidential regression in the regression setting. For volume reconstructed from the heteroscedastic and evidential regression, the corresponding uncertainty fields fail to capture critical features in the error field, indicating the potential limitation of using their predicted uncertainty for error analysis. In contrast, the uncertainty fields produced by CE and CE+MSE show better alignment with the error fields.}
\label{fig:qualitative_error_comparison_between_models}
}

\begin{document}


\firstsection{Introduction}
\maketitle

\input{1-introduction}
\input{2-relative}
\input{3-methodology}
\input{4-experiment}
\input{5-discussion}

\acknowledgments{This research was supported by the U.S. Department of Energy under grants DE-SC-0023320 (Vanderbilt University) and DE-SC-0023319 (University of Arizona). The authors also gratefully acknowledge Matthew Berger for numerous valuable discussions during the preparation of this paper.}

\bibliographystyle{abbrv-doi}
\bibliography{library}
\input{appendix}

\end{document}

%% file: 0-abstract.tex
\abstract{
Implicit neural representations (INRs) offer compact encoding of volumes, but as lossy approximators, inevitably have prediction errors.  We consider INRs that can simultaneously encode relative error scales by predicting distributions using tools from uncertainty estimation.
Typically, uncertainty estimation relies on computationally expensive approaches or on predefined parametric assumptions about the predictive distribution (e.g., Gaussian). 
In this study, we propose a lightweight method that reformulates regression-based INR training as a classification task by discretizing continuous targets into bins, enabling flexible distribution modeling to capture complex multimodal behaviors. 
We analyze the trade-off between regression and classification for INR training and demonstrate that the classification setting tends to achieve high reconstruction quality and competitive error awareness through uncertainty estimation, compared to regression-based approaches.}

%% file: 1-introduction.tex
Implicit neural representations (INRs) have emerged as a powerful tool for compactly representing scientific data, such as 3D volumes and time-varying scalar fields, due to their flexibility and expressiveness.
As approximations of the data, INRs achieve impressive compression ratios~\cite {gu2023nervi,11264349,han2022coordnet,han2025dcinr,han2023kd,lu2021compressive,tang2024ecnr,yang2025meta} and outperform prior state-of-the-art lossy compression methods~\cite{ballester2019tthresh}.
However, their approximating nature introduces errors that can affect the reliability of downstream scientific analysis and raise concerns among domain experts about their use in scientific workflows.

Common error metrics, such as peak signal-to-noise ratio (PSNR), provide a global error estimate for the data approximation but fail to capture localized inaccuracies that are critical for scientific interpretation.  
Without access to the raw data, which is often unavailable
, measuring prediction error is challenging.
Recently, prediction uncertainty has been used as a potential indicator of the relative error scale of INR predictions~\cite{saklani2024uncertainty,xiong2024regularized}.
Classical uncertainty quantification approaches, such as deep 
ensemble~\cite{lakshminarayanan2017simple} and Monte Carlo dropout~\cite{gal2016dropout} provide uncertainty estimates but introduce significant computational overhead or may compromise prediction accuracy.
Lightweight solutions, heteroscedastic~\cite{kendall2017uncertainties,seitzer2022pitfalls} or evidential regression~\cite{amini2020deep}, enable single-pass uncertainty estimation but rely on restrictive distribution assumptions, limiting their ability to model complex scientific data with long-tailed or multi-modal distributions.

In this study, we propose a new lightweight approach for INR-based scientific data modeling that does not rely on predefined parametric assumptions. 
Our solution formulates the regression task as a classification problem by discretizing continuous signals into bins and training with a cross-entropy objective.
Compared with regression-based approaches, the classification formulation yields a discrete predictive distribution that can model complex behavior.

We compare our approach with regression-based methods in terms of reconstruction quality and uncertainty estimation to assess its ability to capture and reflect prediction errors.
Experimental results reveal a trade-off among methods, and our method tends to achieve a high reconstruction quality. 
Furthermore, the error awareness
exhibits a complementary relationship, with each approach performing better under different conditions depending on the dataset properties.
Overall, these findings suggest that classification-based INR training offers a practical and effective alternative for jointly achieving accurate reconstruction and meaningful uncertainty estimation in scientific data modeling. We summarize our contribution as follows:
\begin{itemize}
    \item A new training framework for implicit neural representations that uses the classification objective for implicit neural representation training.

    \item A comparative study showing that the classification approach provides complementary performance in reconstruction quality and uncertainty estimates compared with regression-based solutions.
\end{itemize}

%% file: 2-relative.tex
\section{Related Work}
Prior literature~\cite{han2022coordnet,lu2021compressive,reiser2021kilonerf,saragadam2022miner,tang2023ecnr} on INR studies largely focus on improving compression performance or computational efficiency, while overlooking the characterization of prediction error. Here, we review the prior work on uncertainty estimation in the context of INRs for scientific data modeling.

Model prediction uncertainty is important for ensuring the reliability of neural model decisions~\cite {potter2025navigating}.
Various methods have been proposed to quantify prediction uncertainty in INR training, each motivated by different objectives.
One line of work uses uncertainty as a proxy to indicate whether a region has high or low error.
Saklani et al.~\cite{saklani2024uncertainty} have evaluated the uncertainty of deep ensemble and Monte Carlo dropout over INR for volume visualization tasks. 
Xiong et al.~\cite{xiong2024regularized} proposed RMDSRN, which uses a multi-head decoder to predict variance for uncertainty and evaluate the quality of uncertainty by calculating the Pearson correlation with model prediction error. 
Unlike the deep ensemble, RMDSRN is designed with a shared encoder to reduce the number of model training runs. 
These methods are computationally intense due to the additional model training/inference or introduce additional training complexity.

Another line of work focuses on modeling data that inherently contains uncertainty, such as ensemble simulation outputs. In this setting, it is critical to distinguish between different sources of uncertainty, such as aleatoric and epistemic uncertainty.
ConfEviSurrogate~\cite{duan2025confevisurrogate} builds on the evidential regression~\cite{amini2020deep} to develop a generative surrogate model for scientific simulation. 
REV-INR~\cite{saklani2026rev} integrates evidential and heteroscedastic regression~\cite{kendall2017uncertainties} with a shared-encoder framework~\cite{xiong2024regularized}, resulting in multiple architectures for uncertainty quantification.
Surroflow~\cite{shen2024surroflow} leverages normalizing flows to enable invertible input-output mappings for simulation surrogate modeling and provides uncertainty estimates for predicted output. 

Our study focuses on using uncertainty to characterize prediction errors.
Rather than extending prior approaches, we introduce a lightweight training framework that reformulates INR training using a classification objective.

%% file: 3-methodology.tex
\begin{figure}[t]
\centering   
\includegraphics[width=\linewidth]{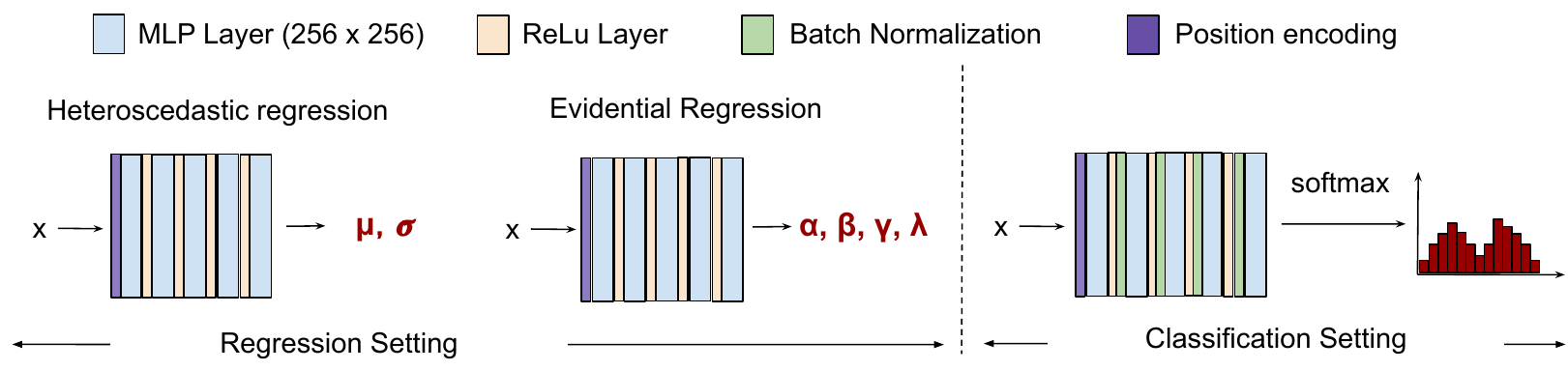}
\caption{The training configuration of heteroscedastic, evidential, and classification.}
\vspace{-3mm}
\label{fig:classification}
\end{figure}

\section{Background}
INRs tend to learn low-frequency features but struggle to capture high-frequency details due to spectral bias~\cite{rahaman2019spectral}.
To address this limitation, techniques like Siren~\cite{sitzmann2020implicit} and position encoding~\cite{muller2022instant,tancik2020fourier} approaches have been proposed to improve model performance. 
As illustrated in Fig.~\ref{fig:classification}, our study adopts Fourier position encoding~\cite{tancik2020fourier} in combination with the ReLU activation across different training settings.
Our experiment uses a 5-layer multi-layer perceptron (MLP)  with a width of 256 neurons per layer.
Beyond standard regression-based INR training, we apply the same framework to model predictive uncertainty using two representative approaches: heteroscedastic and evidential regression.

\textbf{Heteroscedastic Regression}~\cite{kendall2017uncertainties} formulates regression as a Gaussian maximum-likelihood (equation~\ref{eq:heter}) estimation problem.
Given an input $x_i$ and target $y_i$, the model predicts both a mean $\mu_{\theta}(x_i)$ and variance $\sigma^2_{\theta}(x_i)$, where $\theta$ denotes the network parameters.
\update{Instead of minimizing the mean squared error, the negative log likelihood loss jointly optimizes both prediction error and uncertainty($\sigma^2_{\theta}(x_i)$) at the same time}.

\begin{equation}
\mathcal{L}_{Hete\mathrm{NLL}} = -\log(p(y_i\mid x_i)) =\frac{1}{2}\log(\sigma_{\theta}^2(x_i)) + \frac{(y_i - \mu_{\theta}(x_i))^2}{2\sigma_{\theta}^2(x_i)}
\label{eq:heter}
\end{equation}

\textbf{Evidential Regression}~\cite{amini2020deep} models uncertainty by placing a Normal Inverse-Gamma prior over the Gaussian likelihood function. 
The likelihood is $p(y_{i} \mid \mu_{i}, \sigma_{i}^2)=\mathcal{N}(y_{i} \mid \mu_{i}, \sigma_{i}^2)$.
It assumes prior distributions over $\mu_i$ and $\sigma_i$. The $\sigma_i$ follows the Inv-Gamma$(\alpha,\beta)$ distribution and $\mu_i$ follows the gaussian distribution $\mathcal{N}\!\left(\gamma,\,  \frac{\sigma^2}{\lambda}\right)$.

Evidential regression predicts the parameters $\{\alpha_{\theta}(x_i)$, $\beta_{\theta}(x_i)$, $\gamma_{\theta}(x_i)$ $\lambda_{\theta}(x_i)\}$ which define a Student's $t$-distribution. Its negative log-likelihood is:

\begin{equation}
\begin{aligned}
\mathcal{L}_i^{\text{NLL}}(\mathbf{w}) 
&= \frac{1}{2} \log\left(\frac{\pi}{\lambda_{\theta}(x_i)}\right)
- \alpha_{\theta}(x_i) \log\left(\Omega_{\theta}(x_i)\right) \\
&\quad + \left(\alpha_{\theta}(x_i) + \frac{1}{2}\right)
\log\left((y_i - \gamma_{\theta}(x_i))^2 \lambda_{\theta}(x_i) + \Omega_{\theta}(x_i)\right) \\
&\quad + \log\left(
\frac{\Gamma(\alpha_{\theta}(x_i))}{
\Gamma\left(\alpha_{\theta}(x_i) + \frac{1}{2}\right)}
\right)
\end{aligned}
\label{eq:evidential}
\end{equation}

where $\Omega_{\theta}(x_i) = 2\,\beta_{\theta}(x_i)\left(1 + \lambda_{\theta}(x_i)\right)$.
In this study, the predicted mean is given by $\gamma_{\theta}(x_i)$, the aleatoric uncertainty by $\frac{\beta_{\theta}(x_i)}{\alpha_{\theta}(x_i)-1}$, and the epistemic uncertainty by $\frac{\beta_{\theta}(x_i)}{\lambda_{\theta}(x_i)(\alpha_{\theta}(x_i)-1)}$. \update{Since distinguishing between uncertainty sources is not the primary focus, we combine aleatoric and epistemic uncertainty into a single measure $\frac{\beta_{\theta}(x_i)\bigl(1+\lambda_{\theta}(x_i)\bigr)}
{\lambda_{\theta}(x_i)\bigl(\alpha_{\theta}(x_i)-1\bigr)}$}.

For high-quality uncertainty estimation, predictions with higher uncertainty should correspond to larger errors. 
To evaluate how well uncertainty reflects error,  we use the AUSE metric:
\[
\mathrm{AUSE} = \int_0^1 \left( E(p) - E^*(p) \right)\, dp
\]
where $E(p)$ denotes the prediction error after removing a fraction $p$ of the most uncertain predictions, and $E^*(p)$ denotes the error after removing a fraction $p$ of the highest-error predictions (oracle). This metric measures how well uncertainty aligns with true prediction error~\cite{lind2024uncertainty}. A lower AUSE indicates better error awareness.

\section{Classification Setting for INR Training}

Using classification to address regression problems has shown promise in several fields, including computer vision~\cite{pintea2023step, rothe2015dex} and reinforcement learning~\cite{lagoudakis2003reinforcement}. 
In this work, we reformulate regression as a classification problem in the INRs setting to model scientific data without restrictive parametric assumptions.

To transform a regression task into a classification problem, the continuous signal must be discretized into bins. We adopt an equal-range binning strategy, in which values are first normalized to the range $[0, 1]$ and then uniformly partitioned into $n$ bins. The bin size is defined as $b_s = \frac{1}{n}$. The $i$-th bin corresponds to the interval $[b_s \times (i - 1), b_s \times i)$, and its representative value is taken as the midpoint $b_i = \frac{b_s \times (i - 1) + b_s \times i}{2}$. We use the midpoint as the bin value because it typically yields better reconstruction quality than using the left or right bin boundaries. In the classification setting, the target label corresponds to the bin containing the ground-truth value. Unless otherwise specified, we use 256 bins by default \update{(Appendix A provides a further investigation on the bin configuration)}.

In standard classification tasks, models predict a single discrete label corresponding to the highest probability. However, in the INR setting for continuous signal modeling, neighboring bins can also receive relatively high probabilities. As a result, small deviations into adjacent bins still correspond to close numerical values, allowing the model to maintain accurate approximations despite minor prediction shifts.
\update{Rather than representing INR outputs using the value of the target bin, we compute the expected value of the predicted distribution, $\sum_{i=1}^{C} b_i p_i$, where $p_i$ denotes the probability assigned to bin $i$, and C is the number of bins.}

During training, the cross-entropy loss encourages high probability for the target label and low probability for all other bins. This formulation promotes the learning of a meaningful probability distribution that reflects the underlying ground-truth value. The uncertainty of the classification network is captured by the output distribution after the softmax normalization. We quantify prediction uncertainty by computing the variance of this distribution. \update{Our study mainly focuses on using distribution variance for uncertainty estimation, given that variance quantifies how far a prediction may deviate from its expected value, providing an intuitive measure of error awareness. In contrast, uncertainty measures such as entropy are often less directly interpretable in terms of prediction error (check Appendix C).}

\begin{equation}
\mathrm{Var}(y) = \sum_{i=1}^{n} p_i (b_i - \mu)^2, \mu = \sum_{i=1}^{n} p_i b_i.
\end{equation}

Consistent with prior work, we observe that using cross-entropy loss alone can lead to salt-and-pepper noise~\cite{han2025towards}. To mitigate this issue, we introduce a mean squared error (MSE) regularization term to anchor the expected value of the predicted distribution to the ground truth. \update{This regularization reduces noise, quantization error, improves reconstruction quality, and preserves the uncertainty information captured by the cross-entropy objective} \update{(Appendix B provides a brief overview of the impact of $\lambda$)}:

\begin{equation}
\mathcal{L} = \mathcal{L}{\mathrm{CE}} + \lambda\cdot \mathcal{L}{\mathrm{MSE}}.
\end{equation}

%% file: 4-experiment.tex
\section{Reconstruction Performance Comparison}
To compare reconstruction quality, we evaluate four datasets: two simulation datasets (Rayleigh–Taylor instability and magnetic reconnection) and two CT datasets (Bonsai and aneurism), all obtained from the Open Scientific Visualization Datasets~\cite{klacansky2020open}.
To ensure a fair comparison, we design the experimental setup such that each model uses an MLP with approximately the same number of parameters. In the classification setting, the final layer corresponds to the discretized bins. In the regression setting, an additional layer is introduced to maintain a comparable parameter count, resulting in slightly more parameters than in the classification model.
All models are trained using the Adam optimizer with a learning rate of $10^{-3}$ and a batch size of 20{,}000. Each model is trained for 20{,}000 steps. Experiments are conducted on an NVIDIA GeForce V100 GPU with 16 GB of memory.

\begin{figure}[hbt!]
\centering   
\includegraphics[width=0.75\linewidth]{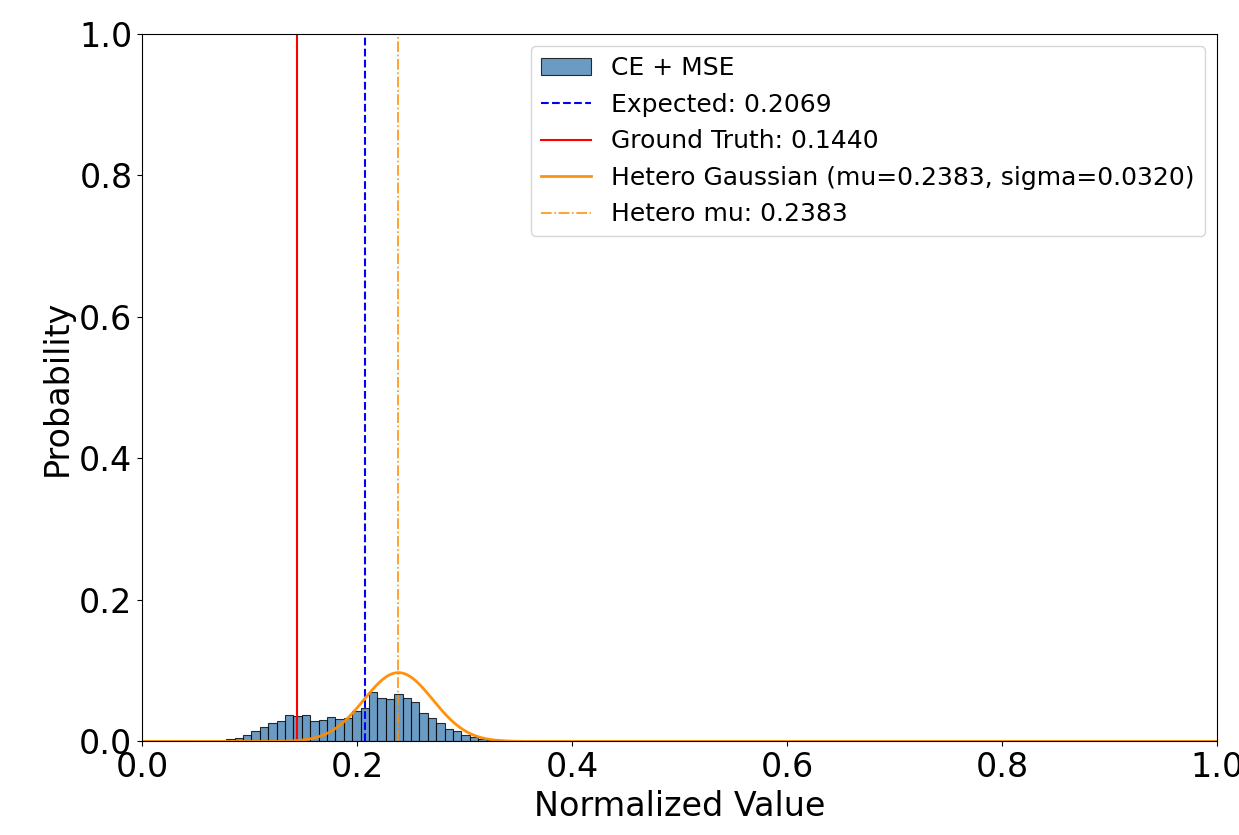}
\caption{Predicted distributions from CE+$\lambda$MSE and heteroscedastic regression. CE+$\lambda$MSE exhibits lower error than the heteroscedastic approach. It predicts a two-peak, multimodal distribution, whereas heteroscedastic regression is constrained by the Gaussian assumption.}
\vspace{-5mm}
\label{fig:classification_multi_modal}
\end{figure}

\subsection{Performance Comparison between Classification and Regression}
Under \textbf{quantitative evaluation} using PSNR, classification-based approaches achieve comparable or even better performance than regression-based methods. The results are summarized in Table~\ref{tab:simulation_data_table_psnr}, which groups methods into classification- and regression-based categories. The table also reports the compression ratio between the original dataset size and the network model size. Here, CE denotes cross-entropy loss, while CE + $\lambda$MSE represents the combination of cross-entropy and mean squared error (MSE) loss. Steel-blue highlights indicate the highest reconstruction quality for each volume, all of which correspond to classification-based (discrete distribution) methods.

Cross-entropy (CE) loss achieves stronger PSNR performance compared with both evidential and heteroscedastic regression. For simulation datasets, CE performs slightly better than regression-based approaches; however, the performance gap becomes much more pronounced for the CT datasets.
On the Bonsai dataset, classification-based methods outperform evidential regression by 2 dB and heteroscedastic regression by approximately 5 dB. A similar trend is observed for the aneurism dataset. As discussed above, combining cross-entropy with MSE further improves performance over using CE alone. In this study, $\lambda = 50$ is selected empirically.
With MSE regularization, the results are generally better than those obtained with pure CE loss. 
In Table~\ref{tab:simulation_data_table_psnr}, we also report the PSNR of MSE under regression-based methods as a reference. 
\update{
MSE-based objectives favor reconstruction accuracy, while CE-based and evidential objectives generally provide more informative uncertainty estimates.
The selection of the preferred method depends on whether the downstream task prioritizes reconstruction accuracy or reliable identification of erroneous regions.}

\begin{figure*}[hbt!]
\centering   
\includegraphics[width=0.9\linewidth]{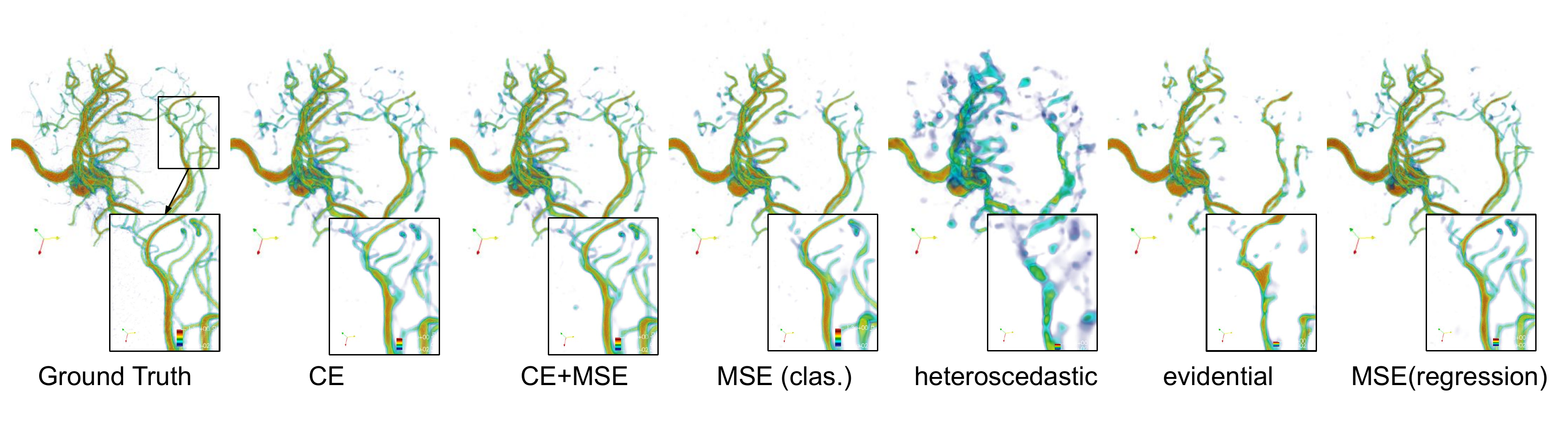}
\vspace{-5mm}
\caption{%
A qualitative comparison between classification-based and regression-based approaches over aneurism dataset. Both heteroscedastic and evidential regression struggle to fit thin vessels and small branches, which are fine-grained and high-frequency features.}
\label{fig:qualitative_comparison_between_models}
\end{figure*}

\begin{table*}[hbt!]
\centering
\caption{Data description of evaluation data and related PSNR value under each training setting. We highlight the best and the second-best performance model. The regular MSE regression solution without uncertainty awareness is used as a reference, but not included for performance comparison.}
\begin{tabular}{cc|ccc|cc|c}
\hline
\multicolumn{2}{c|}{} 
& \multicolumn{3}{c|}{\textbf{Classification-Based}} 
& \multicolumn{3}{c}{\textbf{Regression-Based}} \\

Data & Ratio 
& CE & CE+$\lambda$ MSE & MSE 
& Evidential & Heteroscedastic & MSE\\ 
\hline

Rayleigh-Taylor Instability ($512^3$) & 384X &
31.45 & \colorbox{lightblue}{32.49} & \colorbox{steelblue}{33.84} &
30.51 & 31.03 & 32.35\\

Magnetic Reconnection Simulation ($512^3$) & 384X &
\colorbox{lightblue}{42.90} & 42.88 & \colorbox{steelblue}{43.09} &
42.21 & 41.65 & 43.79 \\ 

bonsai ($256^3$) & 12X &
35.91 & \colorbox{lightblue}{36.03} & \colorbox{steelblue}{36.23} &
33.57 & 30.86 & 36.35\\

aneurism ($256^3$) & 12X &
\colorbox{lightblue}{36.17} & \colorbox{steelblue}{36.89} & 35.46 &
32.70 & 31.76 & 37.08 \\
\hline
\end{tabular}
\label{tab:simulation_data_table_psnr}
\end{table*}

\begin{table*}[hbt!]
\centering
\caption{The AUSE score for error-awareness comparison. A small value is better than a large one.}
\begin{tabular}{cc|ccc|cc}
 \hline
 \multicolumn{2}{c|}{} 
 & \multicolumn{3}{c|}{\textbf{Classification-Based}} 
 & \multicolumn{2}{c}{\textbf{Regression-Based}} \\
Data & Ratio
& CE & CE+$\lambda$ MSE & MSE
& Evidential & Heteroscedastic \\ 
\hline

Rayleigh-Taylor Instability ($512^3$)
& 384x & \colorbox{steelblue}{$2.21\times 10^{-3}$} & \colorbox{lightblue}{$2.29\times 10^{-3}$} & $5.75\times10^{-3}$
& $2.89\times 10^{-3}$ & $3.58\times 10^{-3}$ \\

Magnetic Reconnection Simulation ($512^3$)
& 384x & $2.46 \times10^{-4}$ & \colorbox{lightblue}{$2.45 \times 10^{-4}$} & $2.72\times10^{-4}$
& \colorbox{steelblue}{$1.91 \times 10^{-4}$} & $7.95\times 10^{-4}$ \\

bonsai ($256^3$)
& 12x & \colorbox{steelblue}{$2.61 \times 10^{-4}$} & $9.48 \times 10^{-4}$ & \colorbox{lightblue}{$2.65 \times 10^{-4}$}
& $7.03\times10^{-4}$ & $5.47\times 10^{-4}$ \\

aneurism ($256^3$)
& 12x & \colorbox{steelblue}{$3.44\times 10^{-6}$} & \colorbox{lightblue}{$5.13\times10^{-6}$} & $4.16\times10^{-4}$
& $2.36\times10^{-5}$ & $2.24\times10^{-4}$\\

\hline
\vspace{-6mm}
\end{tabular}
\label{tab:ause_data_table}
\end{table*}

Beyond quantitative analysis, the \textbf{qualitative} results show that INR trained with classification settings enable the reconstruction of high-quality 3D volumes while effectively capturing high-frequency features better than heteroscedastic and evidential regression.
In Fig.~\ref{fig:qualitative_comparison_between_models}, the visualization of the aneurism data from the classification-based methods captures the thin vessel and their relative small branches feature. 
The same features are absent or inadequately captured in heteroscedastic and evidential regression settings. Notably, these fine structures can be captured by MSE-based regression; however, they are not well preserved in heteroscedastic and evidential regression. The uncertainty estimation in these approaches relies on restrictive distributional assumptions (e.g., Gaussian or Student’s $t$-distributions), which constrain their representational flexibility. As a result, these methods introduce a trade-off between uncertainty estimation and prediction accuracy, limiting their ability to model fine-grained features.

To provide a concrete example, Fig.~\ref{fig:classification_multi_modal} shows a predicted output distribution of a single point, which is located at a sharp gradient change region, from the heteroscedastic and the classification setting.
The distribution output from the classification setting is flexible and multi-modal, and has a lower error than the heteroscedastic regression. Instead, heteroscedastic regression is forced to model the output as a Gaussian distribution, which only captures one peak of the multi-modal distribution.


\section{Error-Awareness Evaluation}

To evaluate prediction uncertainty in terms of error awareness, classification-based methods exhibit performance comparable to regression-based approaches. As shown in Table~\ref{tab:ause_data_table}, the lowest AUSE scores are achieved by cross-entropy (CE) and evidential regression methods. Classification-based approaches account for most of the low AUSE scores, except for the Magnetic Reconnection simulation dataset, where evidential regression outperforms the other methods.

In a classification-based setting, the performance trends differ from those observed under PSNR-based evaluation. 
Models trained with the cross-entropy loss typically achieve the lowest AUSE score, where the MSE loss often leads to the highest, and the CE+$\lambda$ MSE yields intermediate performance. 
The MSE loss does not explicitly optimize the output distribution to assign meaningful probabilities; instead, it primarily focuses on minimizing mean squared error.
Unsurprisingly, it results in lower-quality uncertainty estimates compared with the cross-entropy loss function. 
A similar effect can be observed from the $CE+\lambda MSE$ loss, where the uncertainty estimation degrades because of the $MSE$ regularization term.
In the regression-based approach, evidential regression outperforms heteroscedastic on simulation datasets but underperforms in the CT dataset.

AUSE score provides a statistical summary of the quality of uncertainty estimation, but it is critical to assess whether the predicted uncertainty fields help identify error regions in the predicted volume. 
To verify this, we compare the error and uncertainty fields of two training settings using the aneurism dataset. 
Since uncertainty indicates the region with high or low error rather than an absolute error scale, we normalize the error and uncertainty field to the range [0,1] for comparison.
The results in  Fig.~\ref{fig:qualitative_error_comparison_between_models} show a qualitative comparison across different training settings. With the same transfer function, the visualization of the uncertainty fields from heteroscedastic and evidential regressions is sparse and missing fine-grain vessel detail compared with the error field. 

%% file: 5-discussion.tex
\section{Conclusion and Future Works}
In this study, we introduce a lightweight implicit neural representation for scientific data modeling. Our approach reformulates INR regression training as a classification task, enabling high-quality data reconstruction and effective uncertainty estimation for error awareness across multiple datasets. In future work, we will investigate why the uncertainty framework here captures error awareness, in particular, since the data is not inherently uncertain, and explore using classification-based models to train INRs in settings where the sources of uncertainty are different, e.g., ensembles.

%% file: appendix.tex
\clearpage
\appendix
\section*{Appendix}
\addcontentsline{toc}{section}{Appendix}

\begin{figure}[t]
\centering   
\includegraphics[width=\linewidth]{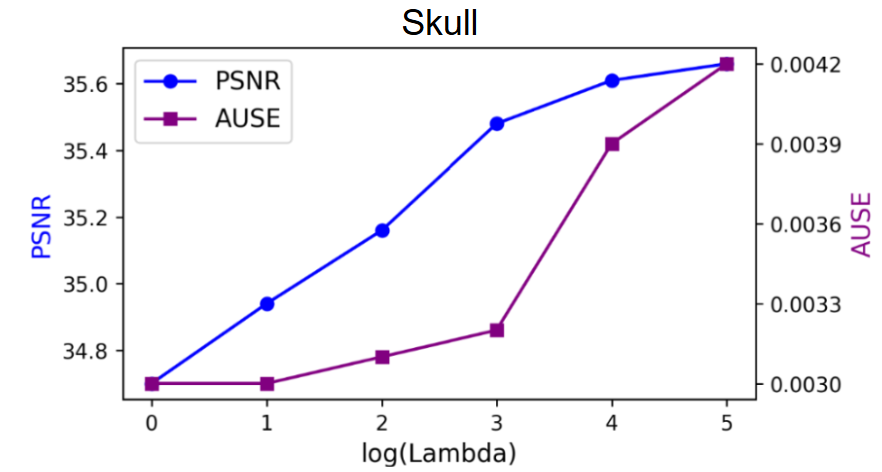}
\includegraphics[width=\linewidth]{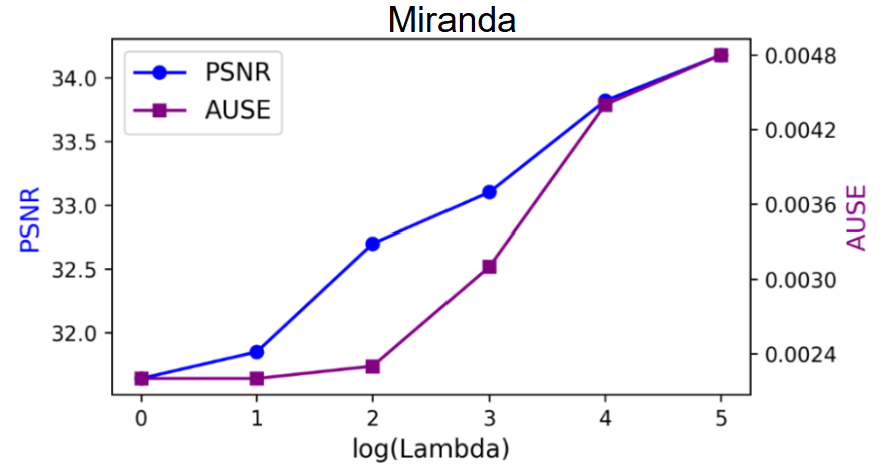}
\caption{The selection of the lambda parameter balances the accuracy of reconstruction and uncertainty quality. Increasing the lambda parameter will improve the reconstruction accuracy (PSNR); however, exceeding a certain threshold decreases the uncertainty quality and increases the AUSE score.}
\vspace{-5mm}
\label{fig:the_selection_of_lambda_parameter}
\end{figure}

\begin{figure}[t]
\centering   
\includegraphics[width=0.98\linewidth]{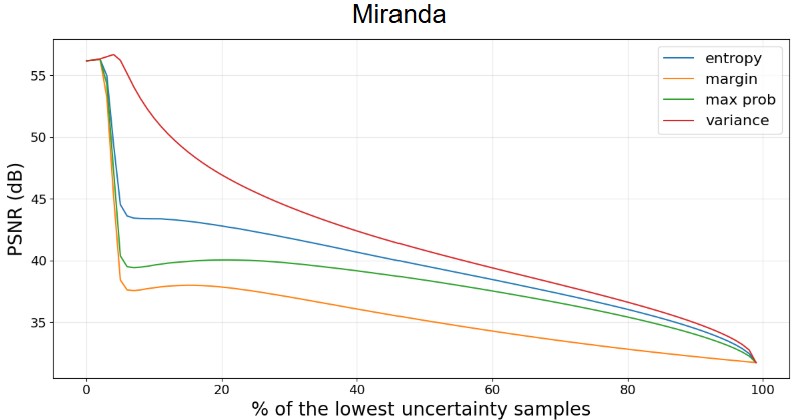}
\includegraphics[width=0.95\linewidth]{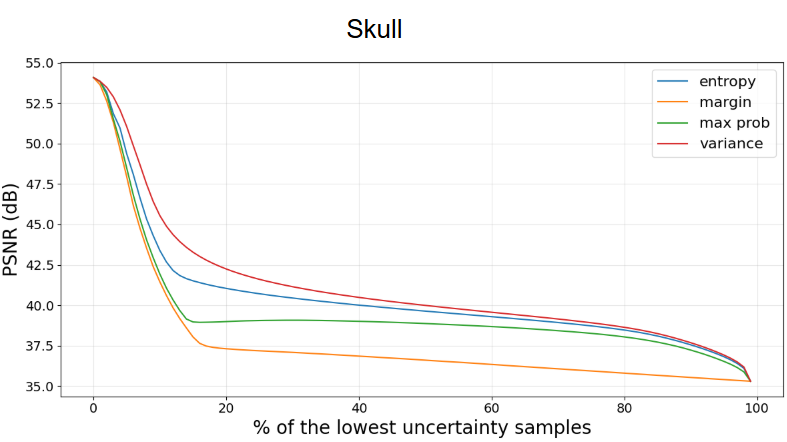}
\caption{PSNR trends constructed using entropy, variance, maximum softmax probability, and margin probability. Across both datasets, the PSNR curve based on variance consistently lies above that based on entropy and other uncertainty metrics, indicating a stronger correlation with prediction error.}
\vspace{-5mm}
\label{fig:psnr_trend_with_lowest_uncertainty_component}
\end{figure}

\section{The configuration of Bin Size for Classification Training}

\subsection{Use midpoint to represent bin value}
\textbf{Number of Bins} affects the reconstruction quality of model predictions. In this study, we use the midpoint representation (Table.~\ref{table:single_multi_prediction}) for bin values, as it yields high reconstruction quality. 

\begin{table}[hbt!]
\centering
\caption{PSNR performance for different choices of representative values per bin (256).}
\begin{tabular}{ccccccc}
 \hline
Simulation Data & left & middle & right\\ \hline

Rayleigh-Taylor Instability  & 31.44 & 31.42 & \colorbox{steelblue}{\textbf{31.46}} \\ 
Magnetic Reconnection Simulation  & 42.73 & \colorbox{steelblue}{\textbf{42.92}}  & 42.43 \\ bonsai & 35.79 & \colorbox{steelblue}{\textbf{35.92}} &  35.74 \\ 
aneurism & 36.13 & \colorbox{steelblue}{\textbf{36.14}} & 35.97 \\
\hline
\end{tabular}
\label{table:single_multi_prediction}
\end{table}

\subsection{How does bin size affect the training performance}
\begin{figure*}[t]
\centering   
\includegraphics[width=0.78\linewidth]{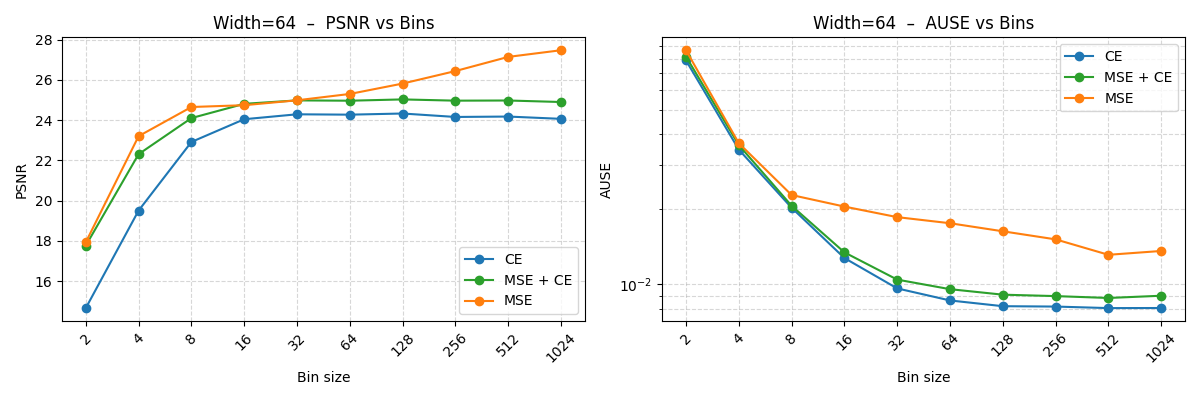}
\includegraphics[width=0.78\linewidth]{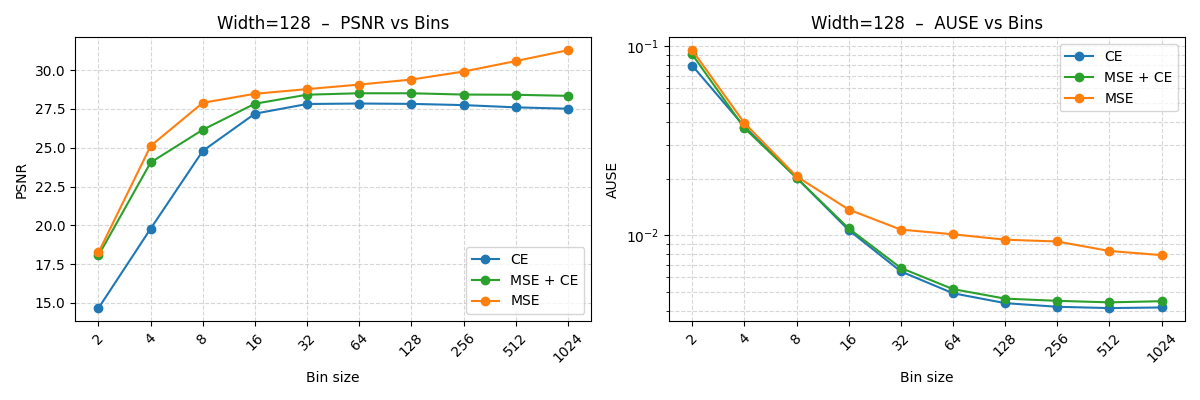}
\includegraphics[width=0.78\linewidth]{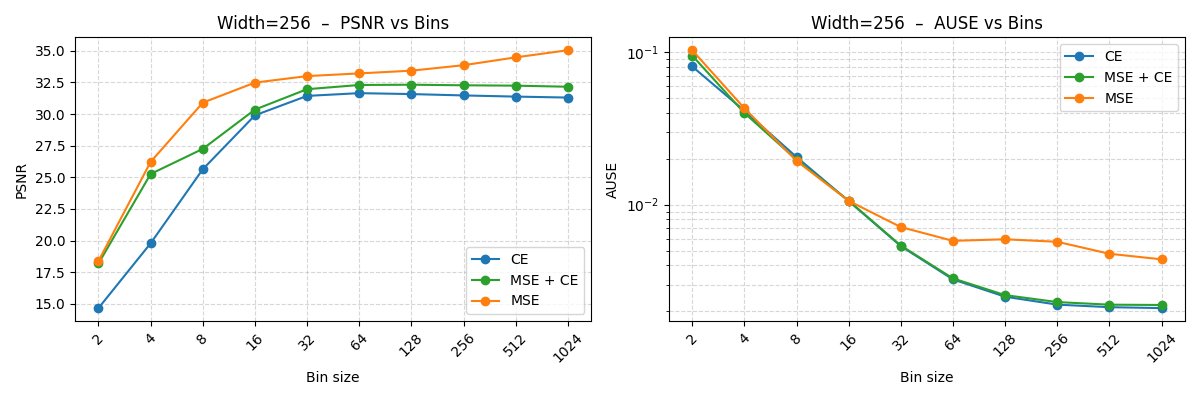}
\includegraphics[width=0.78\linewidth]
{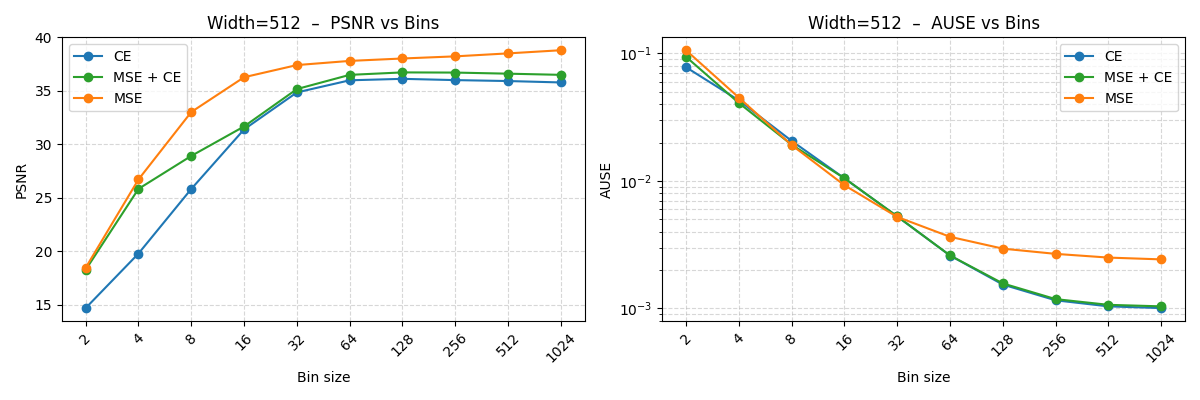}
\includegraphics[width=0.78\linewidth]
{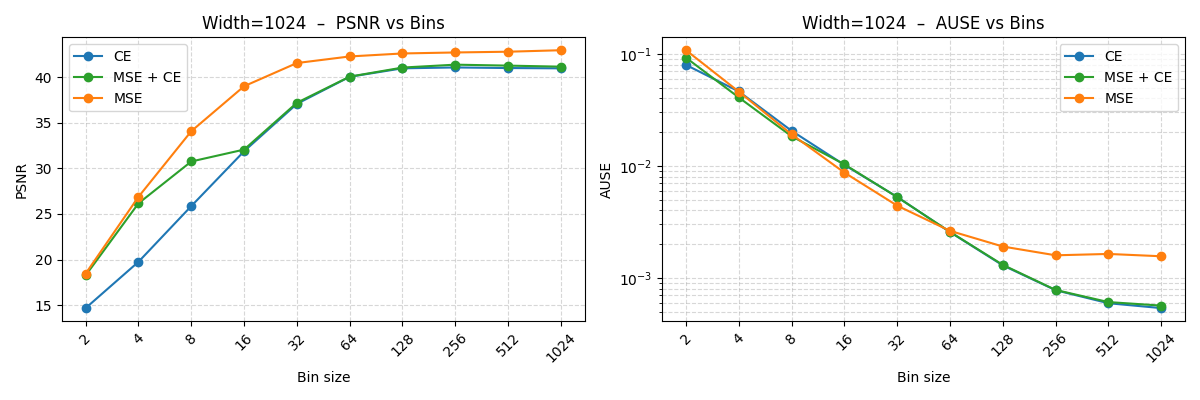}
\caption{The impact of bin size on classification-based network training. The PSNR often plateaus with 128 bins except for the MSE loss. For the AUSE score, a similar trend can be found in all training settings. } 
\label{fig:bin_with_miranda}
\end{figure*}

We present an experiment in which networks with widths of 64, 128, 256, 512, and 1024 are trained using different bin sizes. As shown in Fig.~\ref{fig:bin_with_miranda}, performance plateaus at a bin size of around 128. For classification-based training, performance generally continues to improve as the bin size increases, with the exception of the MSE loss.

\section{The impact of $\lambda$ on accuracy and uncertainty}
To study the effect of the hyperparameter $\lambda$, we evaluate its impact on the Skull and Miranda dataset. Results are shown in Fig.~\ref{fig:the_selection_of_lambda_parameter}.  For relatively small values of $\lambda$, the PSNR score improves while the uncertainty measured by the AUSE score remains unchanged. As the value increases beyond a certainty threshold, the AUSE score begins to deteriorate, indicating a reduction in uncertainty quality. Across both datasets, $\lambda < 100$ has a negligible effect on the uncertainty quality while still providing reconstruction benefits. In our study, we adopt a conservative setting of the $\lambda=50$ throughout our experiments.

\section{Comparing Entropy and Variance for Error Awareness}
We use two datasets to compare the effectiveness of entropy, variance, maximum softmax probability, and margin probability in capturing prediction error under the cross-entropy training framework.
The comparison is based on the PNSR trend, which is generated by progressively adding back the lowest-uncertainty components to the 3D volume and calculating the corresponding PSNR score. In the PSNR equation \ref{eq:psnr}, where $(\mathrm{MAX})$ denotes the maximum possible voxel value. Since all volumes are normalized to the range ([0,1]), ($\mathrm{MAX}=1$), and thus the PSNR is primarily determined by the MSE. Here, $I_i$ and $\hat{I}_i$ denote the ground-truth and predicted voxel values, respectively.

As shown in Fig.~\ref{fig:psnr_trend_with_lowest_uncertainty_component}, the PSNR curves constructed using variance consistently lie above those constructed using entropy for both the Miranda and Skull datasets. This result indicates that variance more effectively identifies components associated with prediction error, leading to faster reconstruction quality improvement as low-uncertainty components are progressively incorporated. Consequently, variance exhibits a stronger correlation with prediction error and serves as a more informative uncertainty metric than entropy and other uncertainty estimation in this setting.

\begin{equation}\label{eq:psnr}
\mathrm{PSNR}
=
10 \log_{10}
\left(
\frac{\mathrm{MAX}^2}{\mathrm{MSE}}
\right),
\end{equation}

\begin{equation}
\mathrm{MSE}
=
\frac{1}{N}
\sum_{i=1}^{N}
\left(
I_i - \hat{I}_i
\right)^2,
\end{equation}
